\documentclass[acmlarge,manuscript,screen]{acmart}

\renewcommand\footnotetextcopyrightpermission[1]{} 
\graphicspath{{figure/pdf/}}

\usepackage{booktabs}
\usepackage{graphicx}
\usepackage{subcaption}
\usepackage[flushleft]{threeparttable}

\usepackage[load-configurations=version-1]{siunitx} 

\newcolumntype{L}[1]{>{\raggedright\let\newline\\\arraybackslash\hspace{0pt}}m{#1}}
\newcolumntype{C}[1]{>{\centering\let\newline\\\arraybackslash\hspace{0pt}}m{#1}}
\newcolumntype{R}[1]{>{\raggedleft\let\newline\\\arraybackslash\hspace{0pt}}m{#1}}
\newcolumntype{N}{@{}m{0pt}@{}}

\AtBeginDocument{%
  \providecommand\BibTeX{{%
    \normalfont B\kern-0.5em{\scshape i\kern-0.25em b}\kern-0.8em\TeX}}}

\begin{document}

\title{Sensor-Based Estimation of Dim Light Melatonin Onset (DLMO) Using Features of Two Time Scales}

\author{Cheng Wan}
\email{chwan@rice.edu}
\affiliation{%
  \institution{Department of Electrical and Computer Engineering, Rice University}
  \city{Houston, TX}
  \country{United States}}

\author{Andrew W. McHill}
\email{mchill@ohsu.edu}
\affiliation{%
  \institution{Oregon Institute of Occupational Health Sciences, Oregon Health \& Science University}
  \city{Portland, OR}
  \country{United States}}

\author{Elizabeth Klerman}
\email{ebklerman@hms.harvard.edu}
\affiliation{%
  \institution{Department of Neurology, Massachusetts General Hospital; Department of Medicine, Brigham and Women's Hospital; Harvard Medical School}
  \city{Boston, MA}
  \country{United States}}

\author{Akane Sano}
\email{akane.sano@rice.edu}
\affiliation{%
  \institution{Department of Electrical and Computer Engineering, Rice University}
  \city{Houston, TX}
  \country{United States}}


\begin{abstract}
  Circadian rhythms influence multiple essential biological activities including sleep, performance, and mood. The dim light melatonin onset (DLMO) is the gold standard for measuring human circadian phase (i.e., timing). The collection of DLMO is expensive and time-consuming since multiple saliva or blood samples are required overnight in special conditions, and the samples must then be assayed for melatonin. Recently, several computational approaches have been designed for estimating DLMO. These methods collect daily sampled data (e.g., sleep onset/offset times) or frequently sampled data (e.g., light exposure/skin temperature/physical activity collected every minute) to train learning models for estimating DLMO. One limitation of these studies is that they only leverage one time-scale data. We propose a two-step framework for estimating DLMO using data from both time scales. The first step summarizes data from before the current day, while the second step combines this summary with frequently sampled data of the current day. We evaluate three moving average models that input sleep timing data as the first step and use recurrent neural network models as the second step. The results using data from 207 undergraduates show that our two-step model with two time-scale features has statistically significantly lower root-mean-square errors than models that use either daily sampled data or frequently sampled data.
\end{abstract}

\ccsdesc[500]{Human-centered computing~Ubiquitous and mobile computing}
\ccsdesc[500]{Applied computing~Health informatics}
\ccsdesc[300]{Computing methodologies~Supervised learning by regression}

\keywords{circadian rhythm, dim light melatonin onset, sensor data, machine learning}

\maketitle

\section{Introduction}
Most human biology and behaviors are heavily influenced by an endogenous circadian oscillator with a period of around 24 hours \citep{czeisler2017human}. Misalignment of this endogenous circadian oscillator with the external environment, which occurs during jet lag and shift work \citep{czeisler1990exposure,sack2007circadian}, negatively affects human health, including increased risk for physical and psychiatric disorders \citep{baron2014circadian}, obesity \citep{di2003effect} and increased 24-hour blood pressure and inflammatory markers \citep{morris2016circadian}. Measuring human circadian phase (i.e., timing) accurately is essential for the clinical treatment of circadian misalignment, possibly improving the efficiency and alertness of shift workers, and for designing other circadian-based interventions. Since the central circadian oscillator for mammals is located in the suprachiasmatic nucleus of the hypothalamus \citep{rusak1979neural}, it is not feasible to measure its status directly in humans. Therefore, markers, including melatonin \citep{pandi2007dim,benloucif2008measuring}, core body temperature (CBT), and cortisol, have been utilized for research and clinical purposes \citep{klerman2002comparisons}. Melatonin-based assessment is the least variable marker among them when assessed under dim light conditions \citep{klerman2002comparisons}. The secretion of melatonin is regulated by various factors including the circadian clock, lighting conditions, and exercise \citep{czeisler2017human}. Under dim light conditions in normally entrained humans, the secretion of melatonin remains at a low level during the day time and increases sharply about 2 hours prior to habitual bedtime \citep{benloucif2008measuring,czeisler2017human}. The time when this increase begins is called dim light melatonin onset (DLMO).

Monitoring melatonin for DLMO, however, requires frequent collection of saliva or blood over at least 7 hours in dim light conditions; this is expensive and inconvenient. Since these samples must be sent for assay, results are not available immediately. Several semi-invasive or non-invasive approaches have been proposed for estimating DLMO using other data types. With the rapid development of wearable devices in the past few years, sensors have been used to collect information about light exposure (LE), skin temperature (ST), and accelerometer (AC) to attempt to predict the timing of sleep onset and sleep offset \citep{sano2018multimodal}. Some work has been conducted using these sensor data for estimating DLMO with the help of machine learning or statistical regression models. Most studies leverage either daily sampled data (sleep onset/offset time) \citep{martin2002sleep,burgess2005dim} or frequently sampled sensor data (including LE, ST, AC every minute) \citep{kolodyazhniy2011estimation,kolodyazhniy2012improved,gil2013human,stone2019generalizability}. Daily sampled data such as sleep timing on previous days contain little information of current day's event. Frequently sampled sensor data can contain many missing values. The combination of the two is likely to provide a better estimation. \citet{bonmati2014circadian} designed several composite phase indexes by simply averaging the onset time or the offset time of various daily variables including LE, ST, body position, and motor activity, and calculated the linear correlations between each phase index and DLMO. For example, their best composite index SleepWT$_\text{On}$ is defined by averaging the sleep onset time and the wrist temperature onset. This method, however, only considers at most two variables and does not include light exposure. Some methods for predicting circadian metrics use limit-cycle oscillators to mathematically describe the dynamics of the circadian oscillators based on frequently sampled LE or sleep-wake cycle data \citep{kronauer1999quantifying,hilaire2007addition,st2007physiologically}. Using the model in \citep{st2007physiologically}, less than 50\% of predictions were within $\pm$1 hour of the observed DLMO in a data set with a wide range of DLMO \citep{phillips2017irregular}.

\begin{figure}[tbp]
  \centering
  \begin{subfigure}[b]{0.95\textwidth}
    \includegraphics[width=\textwidth]{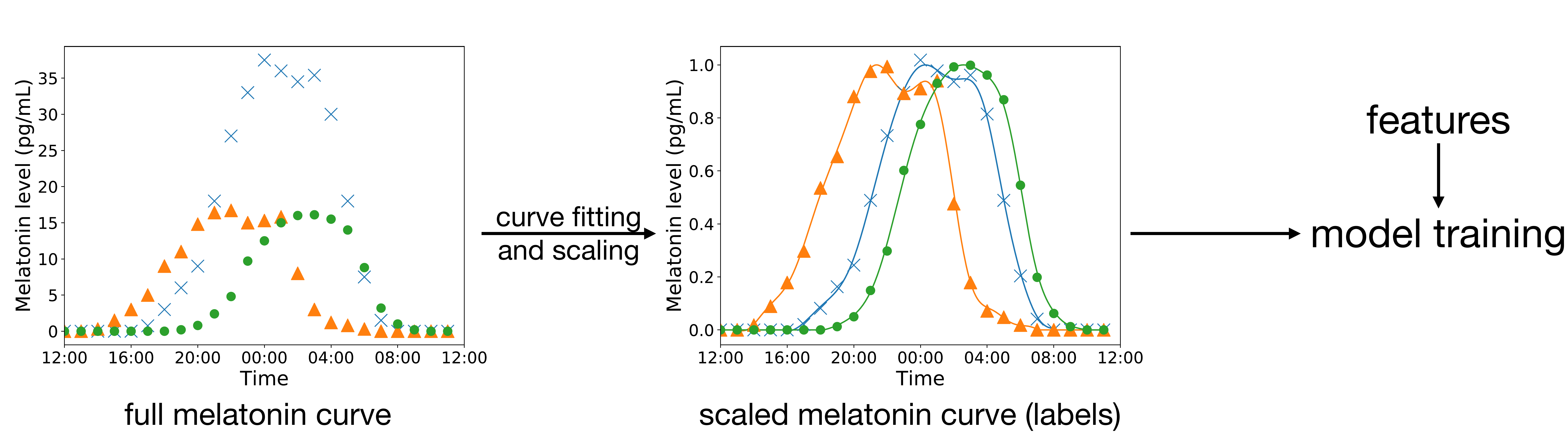}
    \caption{}
    \label{fig:pip1}
  \end{subfigure}
  \begin{subfigure}[b]{0.95\textwidth}
    \includegraphics[width=\textwidth]{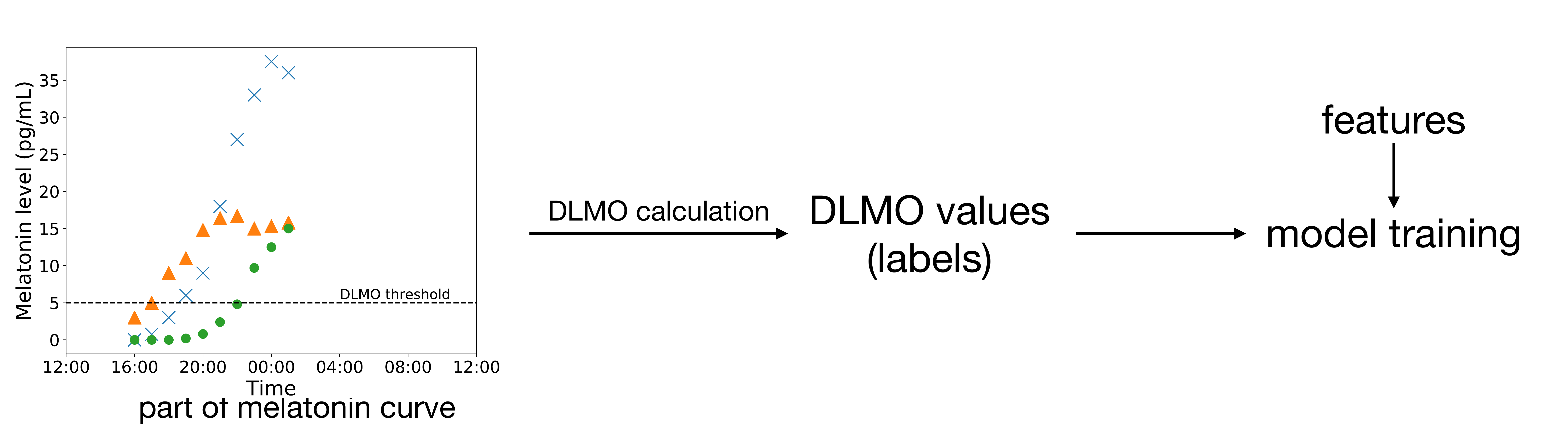}
    \caption{}
    \label{fig:pip2}
  \end{subfigure}
  \caption{A comparison of model training pipelines between previous methods that use frequently sampled data and our method. (a) The pipeline of existing methods that train a model for estimating melatonin level. The method requires fitting and scaling the melatonin curve first, which requires complete melatonin curves. (b) The pipeline of our method that trains a model for estimating the dim-light melatonin onset (DLMO) directly. Different colors denote melatonin profiles of different people. Our method does not require collecting a full melatonin profile.}
  \label{pip}
\end{figure}

In addition to the above mentioned drawbacks, the previous DLMO modeling methods based on frequently sampled sensor data \citep{kolodyazhniy2011estimation,kolodyazhniy2012improved,gil2013human,stone2019generalizability} are designed for estimating melatonin concentrations first, and DLMO values are then calculated based on the predicted melatonin curve. The pipeline of these methods is shown in Figure \ref{fig:pip1}. Since there is a wide inter-individual variation in the peak value of DLMO \cite{burgess2008individual}, these methods need to fit and scale the melatonin curves, which requires the full melatonin profile. This training approach of DLMO that requires the entire melatonin profile (rather than only DLMO), however, greatly increases the cost and participant burdens of melatonin collection because 1) the participants must remain awake for many hours after their habitual bedtime to collect samples and 2) processing and assaying the samples to obtain melatonin measurement is expensive.

In light of these limitations from previous methods, we propose a two-step framework for estimating DLMO using the data from two time scales. In the first step, the framework summarizes all features prior to the day of interest. The second step combines the result of the first step and the current day's frequently sampled data to estimate DLMO. For the implementation, we evaluate three moving average models to summarize sleep onset/offset time for the first step. For the second step, we apply recurrent neural network (RNN) methods to predict DLMO.

The main contributions of this paper are:
\begin{enumerate}
  \item We construct a two-step framework for estimating DLMO using features of two time scales: both daily sampled data and frequently sampled data. This is a generalization of all current models.
  
  \item To implement the framework, we compare three moving average models for the daily sampled data, and evaluate them for extracting features from the daily sleep timing data.
  
  \item To the best of our knowledge, this work is the first to predict DLMO directly using frequently sampled data (using the pipeline in Figure \ref{fig:pip2}), which requires less data collection for training the model.

  \item We show that the model using features of two time scales is significantly better than the one using only one type of feature.

\end{enumerate}

\section{Data Preparation}

\begin{figure}[tbp]
\centering
\centering
\includegraphics[width=0.8\linewidth]{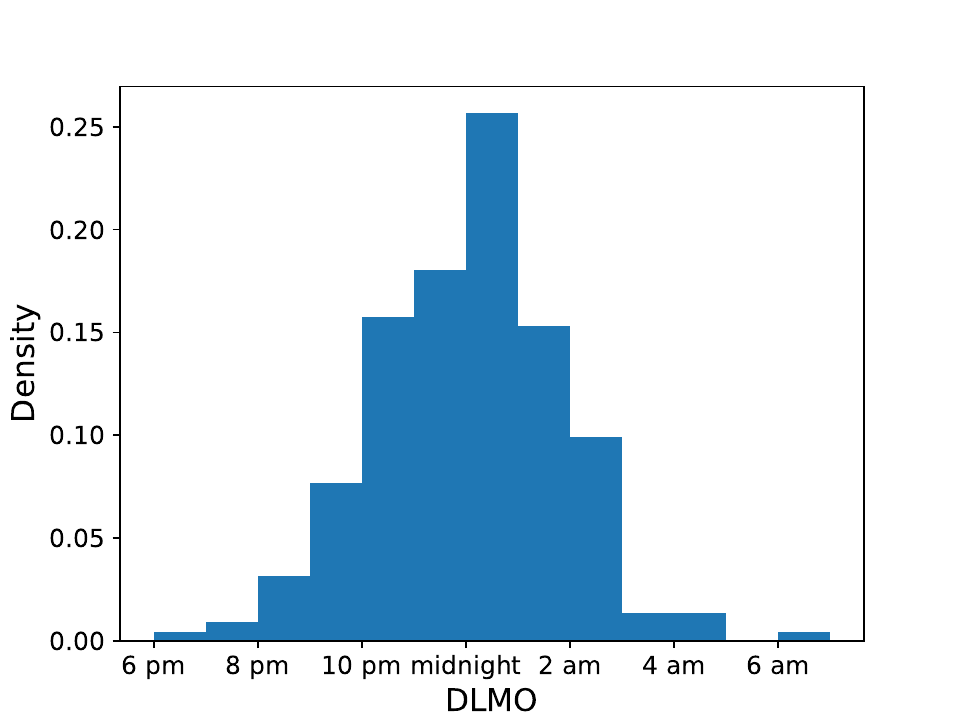}
\caption{The distribution of DLMO values in our dataset. The mean DLMO value is 23.43$\pm$1.76.}
\label{fig:dlmo}
\end{figure}

The data were collected in the SNAPSHOT (\textbf{S}leep, \textbf{N}etworks, \textbf{A}ffect, \textbf{P}erformance, \textbf{S}tress, and \textbf{H}ealth using \textbf{O}bjective \textbf{T}echniques) study, including approximately 1-3 months of daily physiological and behavioral data and one to three DLMO laboratory assessments from 207 undergraduate college students from 2013 to 2017 \citep{sano2016measuring,phillips2017irregular,mchill2017later,taylor2017personalized,sano2018identifying, sano2018multimodal}. Participants in the 2017 cohort had DLMO calculated one to three times over $\sim$3 months of data collection; all other participants have only one DLMO assessment in $\sim$1 month of data collection. We split the data by using the data collected during 2013-2016 ($\sim$1 month of data) as the training set and the data collected in the 2017 cohort ($\sim$3 months of data) as the test set. There are 192 training samples and 31 test samples. Participants were recruited through email.

\subsection{DLMO}
The melatonin concentrations were assayed from saliva collected every hour from 3 pm to 7 am under dim-light conditions ($<$4 lux) \citep{mchill2017later}. The environment was sound-attenuated and temperature-controlled. Participants were permitted to take brief naps between samples and interact with other participants, but were not allowed to use any electronic devices due to the light emitted by such devices \cite{chang2015evening}. Food that might influence the concentration of melatonin in saliva was not allowed during this time. They were also instructed to avoid eating or drinking and to maintain stable posture, for the 20 min prior to each sample. DLMO was determined by linear interpolation of the time that melatonin values first exceeded a threshold of 5 pg/mL \citep{phillips2017irregular}. The distribution of DLMO values collected in our dataset is shown in Figure \ref{fig:dlmo}. The mean DLMO value is 23.43$\pm$1.76.

\subsection{Physiological and Behavioral Data}
During the experiments, the participants wore (i) a wrist sensor on their dominant hand (Q-sensor, Affectiva, USA) to continuously measure ST and three-axis AC at 8Hz sampling rate, and (ii) a wrist actigraphy monitor on their non-dominant hand (Motion Logger, AMI, USA) to measure LE stored at 1 minute intervals. We logarithmically transformed the LE values and computed the L2-norm for AC. We segmented the data into one-hour bins and computed the mean values over each hour as the features of that hour. Sleep onset/offset times were computed based on sleep diary queries every morning and Motion Logger data analyzed using Action-4 software \citep{sano2018multimodal}. For any missing data of each feature, we used linear interpolation to impute the data for missing data less than 12 hours, which performs the best for our dataset with other imputation methods (data not shown). We did not use data sets that contain missing segments longer than 12 hours for the day prior to DLMO collection. After those data sets were removed, the longest missing segment was about 8 hours.

\subsection{Removing the Effect of Daylight-Saving Time}
Since some participants were studied during the clock shift between standard time and daylight saving time, we subtracted one hour from the clock time during daylight-saving time to make clock times uniform across all participants. This standardization is essential for the following reasons:
\begin{enumerate}
    \item Our target is to design a model that predicts DLMO tracking the user activity. A time in our dataset is a feature (sleep time) or a label (DLMO). By standardizing the times, the model does not need to assess whether the clock time is changed.
    \item If the clock time changes, user activity would also change due to social constraints. The standardization of time clock allows the model to capture this adjustment, which could provide better prediction performance intuitively.
    \item Some participants in our dataset experienced the clock time change while some did not. This standardization makes the model perform consistently for both groups of participants.
\end{enumerate}

\section{Methods}

We represent human circadian phase by the following formula
\begin{equation}
  \phi_t=f(\psi_{t-1},x_t)\label{fml:dlmo0}
\end{equation}
where $\phi_t$ is the circadian phase (e.g., DLMO) on day $t$, $\psi_{t-1}$ is the extracted features from the data prior to the day $t$, $x_t$ is the 24-hour frequently sampled data starting from the wake-up time of the $t$ th day, and $f(\cdot, \cdot)$ is the model that combines all features for estimating $\phi_t$. Previous studies have described this relationship mathematically when $x_t$ is LE or sleep timing data, and $\psi_{t-1}:=\phi_{t-1}$ \citep{kronauer1999quantifying,hilaire2007addition}. In practice, the true value of $\phi_{t-1}$ is never known. In this section, we introduce our methods of estimating DLMO under this framework. Since we are only interested in estimating the DLMO values on a specific day, $t$ in Formula \ref{fml:dlmo0} is constant, thus, we rewrite this formula as follows.
\begin{equation}
    \phi=f(\psi,x)\label{fml:dlmo1}
\end{equation}

The framework is composed of the following two steps:
\begin{enumerate}
    \item Extracting features from the data prior to the current day, (i.e., calculating $\psi$).
    \item Estimating DLMO with $\psi$ and current day's frequently sampled data $x$ (e.g., LE, ST, and AC) (i.e., modeling $f$).
\end{enumerate}

It should be noted that while $x$ only consists of frequently sampled data of the current day, $\psi$ can be derived from both daily sampled data and frequently sampled data before the current day. This framework is the extension of the models that use only one type of time-scale data.

\begin{itemize}
  \item For models only using daily sleep onset/offset times \citep{martin2002sleep,burgess2005dim}, $\phi=\psi$ where $\psi$ is the features derived by the daily data.
  \item For models only using frequently sampled data of the current day \citep{kolodyazhniy2011estimation,gil2013human,stone2019generalizability,kolodyazhniy2012improved}, $\phi=f(0,x)$.
\end{itemize}

\subsection{Our Implementation of the Two-Step Framework}

\label{sec:rnn_ema}
\begin{figure}[tbp]
  \centering 
  \includegraphics[width=1\linewidth]{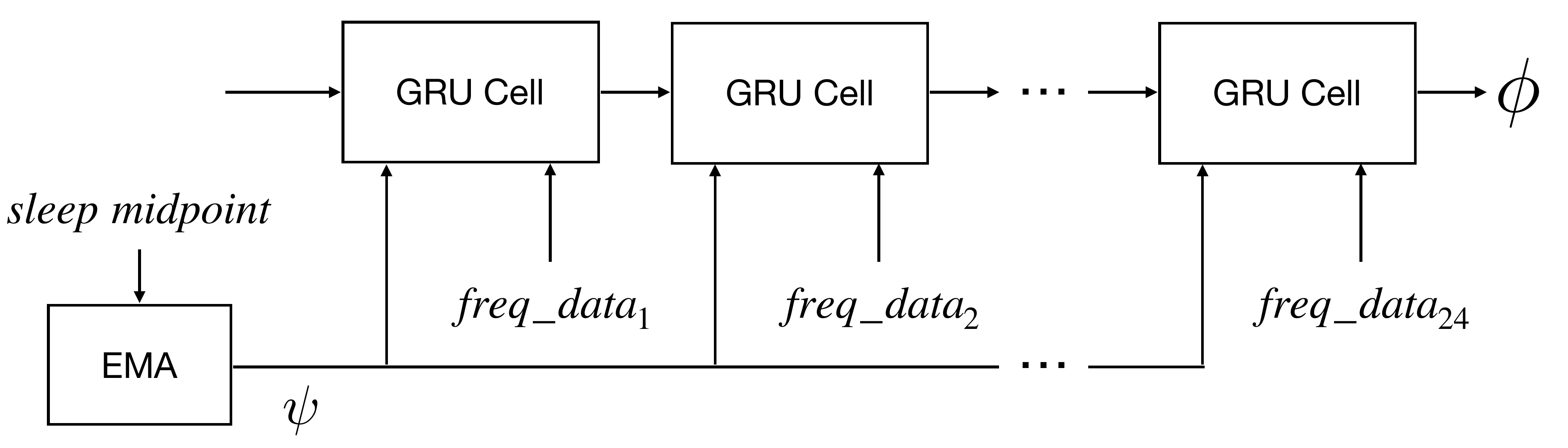} 
  \caption{The structure of RNN$_{\text{EMA}}$ as an example of our framework. In this model, we first use an exponential moving average (EMA) model for calculating $\psi$ from 7-day daily sleep parameters. Then we feed the hourly frequently sampled features ("freq\_data") of light exposure, skin temperature, physical activity, and the output of the EMA model into different RNN cell (gated recurrent unit (GRU) in our implementation) for calculating current day's DLMO $\phi$.}
  \label{fig:model}
\end{figure}

In the first step of the model, we use daily sleep midpoint (i.e., midpoint between sleep onset and sleep offset times) during the past week for deriving $\psi$, and define $\psi$ as an estimator of $\phi$. This definition allows $\psi$ to be a bridge between daily sleep timing parameters and $\phi$. The sleep midpoint of the $i$th day over the past $n$ days is denoted as $T_i$. In this work, we set $n=7$. The first model we evaluate is a simple moving average (SMA) model which was used in \citep{martin2002sleep,burgess2005dim,crowley2006estimating}:

\begin{align}
\psi_{\text{SMA}}&=a\sum_{i=1}^nT_i+b
\end{align}
where $a$ and $b$ are trainable parameters, and are trained by least squares.

The second model for evaluating $\psi$ is an exponential moving average (EMA) model. The model can be presented as

\begin{align}
\psi_{\text{EMA}}&=a\sum_{i=1}^n\alpha^{n-i}T_i+b
\end{align}

where $a$ and $b$ are trainable parameters, and $\alpha$ is a fixed decay rate. We choose 0.9 as the value of $\alpha$ as we found this value was the best parameter for modeling $\phi$ in our preliminary experiments (data not shown).

We also introduce a moving average model (MA, a more generalized model) for comparison:
$$\psi_{\text{MA}}=\sum_{i=1}^nw_iT_i+b$$
where $w_i$ and $b$ are trainable parameters. 

For the second step of the model, we apply a recurrent neural network (RNN) for estimating $\phi$. RNN models have shown great success in natural language processing and sequence learning \citep{haykin1994neural,mikolov2010recurrent}. For the $i$th cell, time, $\psi$ and the frequently sampled data $x$ (i.e., LE, wrist ST, and AC) of the $i$th hour on that day are used as the inputs. We input $\psi$ to each cell because the timing of light stimuli impacts the direction and magnitude of the circadian phase shift \cite{khalsa2003phase}. In our implementation, we use gated recurrent unit (GRU) cells for the RNN model because they have better performance on small datasets \citep{chung2014empirical}. As an example, our structure of RNN$_{\text{EMA}}$ that uses EMA in the first step is shown in Figure \ref{fig:model}.

The training process contains three independent steps. First, we use least squares to train the model that uses sleep midpoints. Then, we fix the parameters in this model and train an RNN. Lastly, we fine-tune all trainable parameters in the model using back-propagation.

\subsection{Comparison of Moving Average Models}
\label{sec:hpt}
\paragraph{SMA v.s. EMA} We consider the following two averages of the sleep midpoint time
\begin{equation}
\overline{T}_t^{\text{SMA}}=\frac{\sum\limits_{i=0}^{n-1}T_{t-i}}{n}\notag
\end{equation}
\begin{equation}
\overline{T}_t^{\text{EMA}}=\frac{\sum\limits_{i=0}^{n-1}\alpha^iT_{t-i}}{\sum\limits_{i=0}^{n-1}\alpha^i}\notag
\end{equation}
Here $\overline{T}_t^{\text{SMA}}$ represents a simple moving average and $\overline{T}_t^{\text{EMA}}$ is an exponential moving average. Suppose $T_i$ follows a Gaussian channel:
$$T_i=\widetilde{T}_i+\varepsilon_i$$
where $\widetilde{T}_i$ is the true value of sleep midpoint with i.i.d. noise $\varepsilon_i\sim\mathcal{N}(0,\sigma^2)$. Therefore,
\begin{equation}
\overline{T}_t^{\text{SMA}}\sim\mathcal{N}\left(\frac{\sum\limits_{i=0}^{n-1}\widetilde{T}_{t-i}}{n},\frac{\sigma^2}{n}\right)\notag
\end{equation}
\begin{equation}
\overline{T}_t^{\text{EMA}}\sim\mathcal{N}\left(\frac{\sum\limits_{i=0}^{n-1}\alpha^i\widetilde{T}_{t-i}}{\sum\limits_{i=0}^{n-1}\alpha^i}, \frac{\sum\limits_{i=0}^{n-1}\alpha^{2i}}{\left(\sum\limits_{i=0}^{n-1}\alpha^i\right)^2}\sigma^2\right)\notag
\end{equation}\\
According to Cauchy-Schwarz inequality, $\left(\sum\limits_{i=0}^{n-1}\alpha^i\right)^2\leq n\sum\limits_{i=0}^{n-1}\alpha^{2i}$. $\overline{T}_t^{\text{EMA}}$ has a larger variance than $\overline{T}_t^{\text{SMA}}$. When $\sigma$ is large (data are noisy), the average value of the sleep midpoint is a better approach for reducing the noise.\\
Intuitively, $\overline{T}_t^{\text{EMA}}$ is a more reliable feature for estimating DLMO because recent sleep timing would be expected to be more highly correlated with DLMO than distant sleep timing as sleep gates the light exposure that might shift DLMO \citep{czeisler2017human}. In addition, sleep timing of each day is highly influenced by circadian phase of that day as human circadian rhythm strongly regulates human sleep timing \citep{mistlberger2005circadian,czeisler2017human}. Therefore, we hypothesize that the EMA model is more sensitive to noise, and it could be a better method for estimating DLMO when the noise is small.\\

\paragraph{SMA/EMA v.s. MA} MA is an extension of SMA/EMA with more tuneable parameters. Therefore, more data are required for training MA. \citet{green1991many} suggests a sample size of $N\geq50+8m$ for the multiple regression where $m$ is the number of predictor variables. 

\section{Experiments}
For the two-step model we designed, we tested three questions:

\begin{enumerate}
    \item In the first step, which moving average model performs the best for estimating DLMO?
    \item Does the two-step model perform better for estimating DLMO than models that only use either daily sampled or frequently sampled data?
    \item  Which combination of frequently sampled data (ST, AC, and LE) performs the best for estimating DLMO? Fewer types of data reduce the cost of required sensors.
\end{enumerate}

We designed three experiments to address these three questions. The code was implemented in Python and the machine learning framework we used was PyTorch \cite{NEURIPS2019_9015}.

\subsection{Question 1: Comparison among Moving Average Models}
\label{sec:exp1}

We first compare the performance of three moving average models: SMA, EMA and MA.

\begin{enumerate}
\item We calculated three values for comparing performance:
\begin{itemize}
  \item RMSE$_\text{all}$: We calculate the root-mean-square error (RMSE) using all data for training the linear regression model.
  \item RMSE$_\text{val}$: We evaluate the model's generalization ability for avoiding overfitting problems, by calculating RMSE on a leave-one participant-out cross-validation data set.
  \item $<$1h: We calculate the percentage of the test samples with absolute error of less than one hour using a leave-one participant-out cross-validation data set. Similar metrics are also reported in \cite{burgess2005dim,phillips2017irregular}.
\end{itemize}

\item We evaluated the effect of statistical noise in the data on the SMA, EMA, and MA models by adding Gaussian noise with standard deviation $\sigma$ to each value of the sleep midpoint. For each $\sigma$, we repeatedly generated noisy sleep data and calculated the coefficient of determination ($r^2$) between true DLMO and the predicted DLMO using noisy sleep data 20,000 times.

\item We also tested model performance with different window sizes between 3 and 8 days (i.e., the number of days of data in the model).
\end{enumerate}

\subsection{Question 2: Performance of the Two-Step Model}
\label{sec:exp2}
Based on the first experiment, we choose the best moving average model $M$ for implementing a two-step model. We compare the performance of the model $M$, the corresponding two-step model RNN$_M$, and the RNN model with 24-hour frequently sampled data RNN$_\text{24-hour}$. Here we do not leverage more than 24 hours of frequently sampled data because previous work \citep{kolodyazhniy2011estimation} reports that 24-hour data is better than longer data sets for predicting DLMO, and longer data set are more likely to contain longer missing data segments. In our dataset, at least 25\% of 7-day (168-hour) frequently sampled data contain missing segments longer than 12 hours. The removal of the missing segments leaves smaller amounts of data to analyze, and makes it harder to compare the results from the model with 7-day frequently sampled data with other models.

We use the following metrics to compare the three models.

\begin{itemize}

\item RMSE$_\text{val}$: To reduce the computation cost of evaluating the model, we run 10-fold cross-validation instead of leave-one-out cross-validation on the training set to get the hyperparameters with the lowest RMSE. We define the RMSE value of the $i$th validation set as $r_i$, we define RMSE over all validation sets as RMSE$_\text{val}=\sqrt\frac{r_1^2+r_2^2+\cdots+r_{10}^2}{10}$.

\item RMSE$_\text{test}$: After selecting the best hyper-parameters based on RMSE$_\text{val}$ and training the model using all training samples, we calculate the RMSE value on the test set.

\item $<$1h: We calculate the percentage of the test samples with absolute error of less than one hour.

\end{itemize}

\subsection{Question 3: Comparison among Different Input Combinations}

To address the third question, we compare the models with different combinations of frequently sampled data. For the best model $M$ in the first experiment (Section \ref{sec:exp1}), we replace its frequently sampled data with one type of input or arbitrary two-type-of-input combination. A one-input model with LE, ST or AC is denoted as RNN$_M^{\text{(LE)}}$, RNN$_M^{\text{(ST)}}$ and RNN$_M^{\text{(AC)}}$ respectively. Similarly, we define the models with two types of input as RNN$_M^{\text{(LE, ST)}}$, RNN$_M^{\text{(LE, AC)}}$ and RNN$_M^{\text{(ST,AC)}}$. We use the same metrics as in Section \ref{sec:exp2}.

We use cross validation and Akaike Information Criterion (AIC) at the same time for selecting the best combination of features because all criteria for feature selection from high-dimensional and small sample data have drawbacks. For example, (i) small bias exists in selecting features using cross-validation when the feature number is small compared with data size \citep{smialowski2009pitfalls}. Note that cross validation is always a biased estimate for feature selection as the feature selection process requires accessing test data \citep{refaeilzadeh2007comparison}. (ii) AIC is an information theory-based metric that considers the performance of the model and the number of parameters in the model. Note that there are many assumptions and approximations in AIC \citep{davies2005cross}, and the degrees of freedom (DOF) in neural networks are generally much less than the number of parameters \citep{murata1994network,gao2016degrees}. For our RNN work, DOF has not yet been studied. (iii) Information theory-based feature selection methods cannot be applied in our dataset because our dataset is not large enough for estimating the mutual information between each high dimensional feature and the DLMO value.

\section{Results}

\subsection{Question 1: Comparison among Moving Average Models}
\label{sec:res1}
\begin{figure}[h]
\centering
\includegraphics[width=0.8\linewidth]{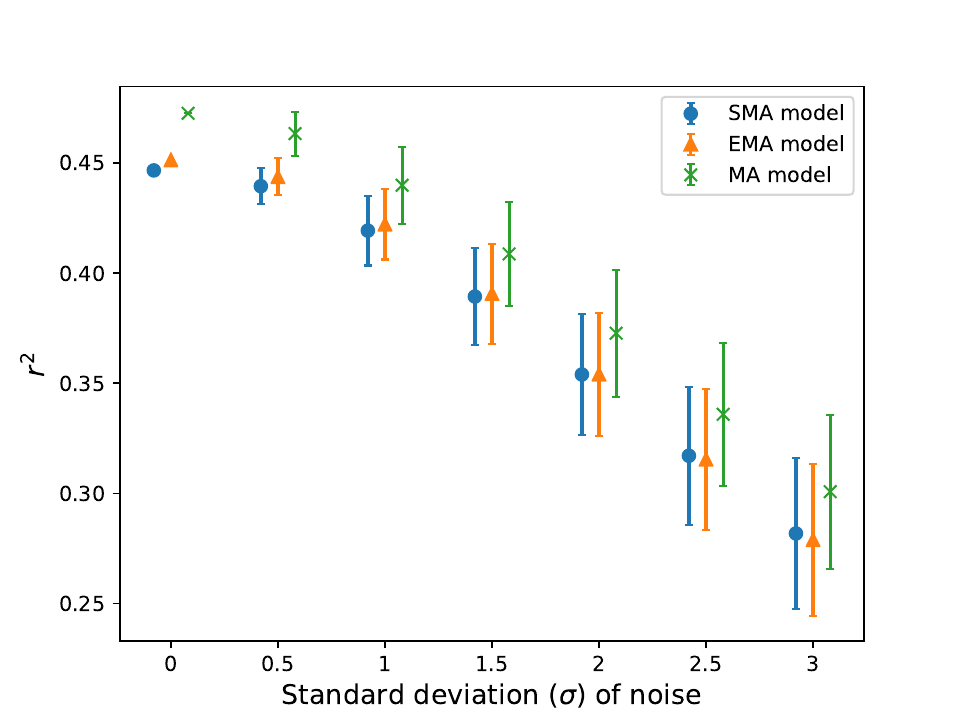}
\caption{The performance of each moving average model with different different  standard deviation of noise.}
\label{fig:noise}
\end{figure}

The three moving average models perform similarly on the validation sets ($p=0.97$, repeated measures ANOVA) (Table \ref{tab:comp1}). For the model performance of moving average models after adding noise (Figure \ref{fig:noise}), the $r^2$ values followed Gaussian distribution for each $\sigma$ value ($p<0.002$, D'Agostino's $K$ test). The MA model showed the highest $r^2$ between DLMO and sleep data (Figure \ref{fig:noise}). When the level of noise was low, the features extracted by the EMA model showed larger $r^2$ than SMA. With increased noise, all models performed worse. SMA performed slightly better than EMA when $\sigma>2$. When $\sigma=2$, there was no significant difference between the $r^2$ of SMA and EMA ($p=1.0$, two-sample t-tests with Bonferroni correction). For other $\sigma$s the differences were significant ($p<0.001$,  statistically but there differences may not be important in practice.).

There was no significant difference for each model by different data window sizes ($n$ in the moving average models) ($p>0.05$, within-group multiple paired comparison).

\begin{table}
\centering 
\begin{threeparttable}
  \caption{A comparison among 3 moving average models.}
  \label{tab:comp1} 
  \begin{tabular}{|C{2cm}|C{2cm}|C{2cm}|C{2cm}|}
      \hline
    Model    & RMSE$_\text{all}$    & RMSE$_\text{val}$    & $<$1h\\
    \hline
    SMA    & 1.36            & 1.38    & 56.5\%\\
    \hline
    EMA        & 1.36            & 1.37    & 56.9\%\\
    \hline
    MA        & 1.33    & 1.38    & 56.8\%\\
    \hline
  \end{tabular}
\end{threeparttable}
\end{table}

\subsection{Question 2: Performance of the Two-Step Model}

Since all three moving average models performed similarly (Section \ref{sec:res1}), we implemented RNN$_\text{SMA}$ (RMSE$_\text{val}$=1.25, RMSE$_\text{test}$=1.34, $<$1h: 61.3\%), RNN$_\text{EMA}$ (RMSE$_\text{val}$=1.21, RMSE$_\text{test}$=1.38, $<$1h: 64.5\%), and RNN$_\text{MA}$ (RMSE$_\text{val}$=1.26, RMSE$_\text{test}$=1.36, $<$1h: 48.4\%). 
The training time for the cross validation sets was about 30 minutes on a CPU (2.6 GHz Intel Core i7). The difference among these models on validation sets was not significant ($p=0.66$, repeated measures ANOVA). Therefore, in the next step, we used the model RNN$_\text{EMA}$ because it had the best score on two of the three metrics (i.e., RMSE$_\text{val}$ and $<$1h).

\begin{figure}[tbp]
\centering
\centering
\includegraphics[width=1.0\linewidth]{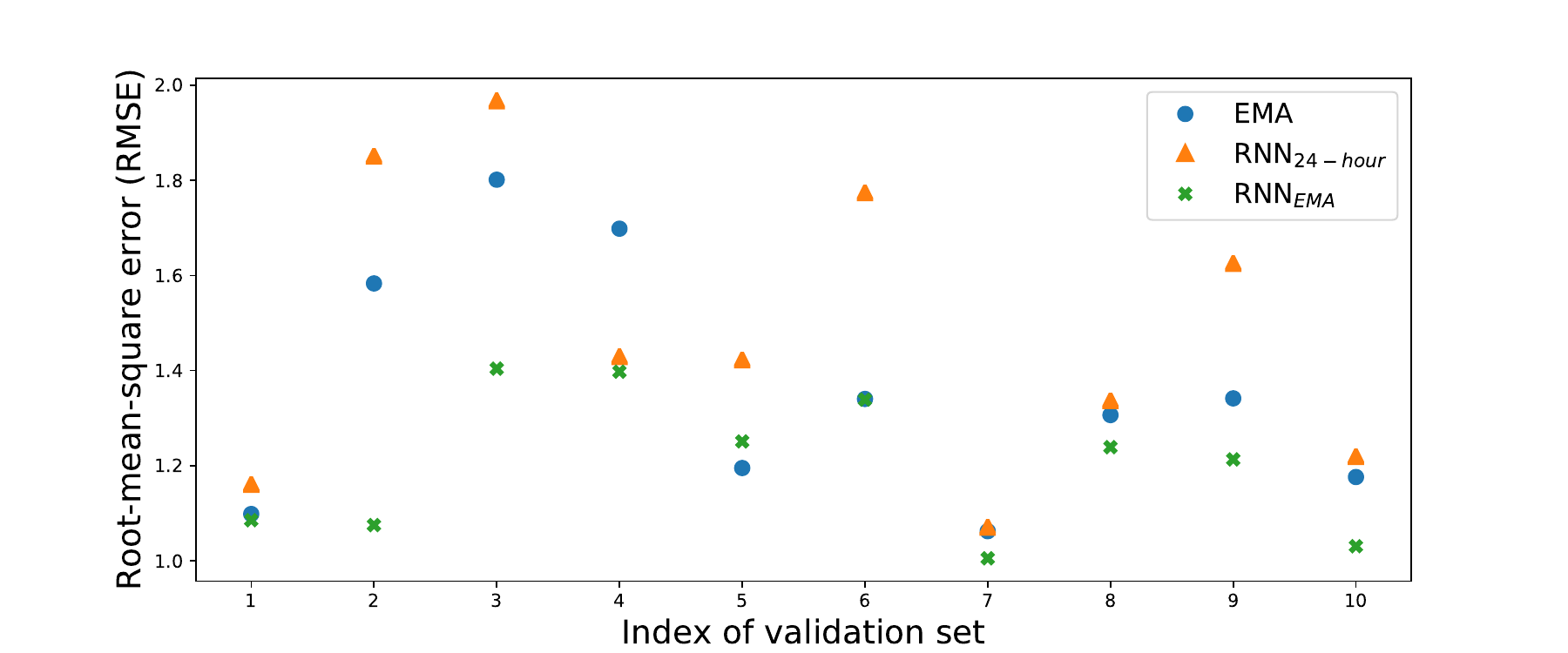}
\caption{The root-mean-square error of each validation set for the three models.}
\label{fig:rmse}
\end{figure}

Table \ref{tab:struct} shows that RNN$_\text{EMA}$ showed the lowest RMSE$_\text{val}$ and RMSE$_\text{test}$ and highest $<$1h. 
The model performance for EMA, RNN$_\text{24-hour}$, and RNN$_\text{EMA}$ on the validation sets (RMSE$_\text{val}$) was significantly different    ($p=0.0022$, repeated measures ANOVA). RNN$_\text{EMA}$ performed significantly better than the other two models (for RNN$_\text{EMA}$ v.s. EMA, $p=0.048$; for RNN$_\text{EMA}$ v.s. RNN$_\text{24-hour}$, $p=0.013$, using multiple paired t-test with Bonferroni correction). Therefore, the combination of 7-day daily sampled data and 24-hour frequently sampled data (RNN$_\text{EMA}$) showed statistically significantly better model performance than EMA or RNN$_\text{24-hour}$ (Figure \ref{fig:rmse}). Note that while the RMSE varied across each validation set, the RMSE for RNN$_\text{EMA}$ was lowest in 9/10 sets.

\begin{table}
\centering 
\begin{threeparttable}
  \caption{A comparison among three different structures.}
  \label{tab:struct} 
  \begin{tabular}{|C{3cm}|C{2.2cm}|C{2.2cm}|C{2cm}|}
      \hline
    Model    & RMSE$_\text{val}$    & RMSE$_\text{test}$& $<$1h\\
    \hline
    EMA                        & 1.38        & 1.42            & 54.8\%    \\
    \hline
    RNN$_\text{24-hour}$    & 1.51        & 1.55    & 41.9\% \\
    \hline
    RNN$_\text{EMA}$        & 1.21    & 1.38    & 64.5\%\\
    \hline
  \end{tabular}
\end{threeparttable}
\end{table}

\subsection{Question 3: Comparison among Different Input Combinations}

\begin{table}
\centering 
\begin{threeparttable}
  \caption{A comparison among models with different input.}
  \label{tab:feature} 
  \begin{tabular}{N|C{3cm}|C{2.2cm}|C{2.2cm}|C{2cm}|C{2cm}|}
      \hline
    & Model    & RMSE$_\text{val}$    & RMSE$_\text{test}$& $<$1h & AIC\\
    \hline
    & RNN$_\text{EMA}$        & 1.21    & 1.38    & 64.5\% & 207.6\\
    \hline
    \rule{0pt}{16pt}& RNN$_\text{EMA}^{\text{(LE)}}$    & 1.31        &  1.44       & 61.3\% & 247.3\\
    \hline
    \rule{0pt}{16pt}& RNN$_\text{EMA}^{\text{(ST)}}$    & 1.27        & 1.48        & 45.2\% & 216.6\\
    \hline
    \rule{0pt}{16pt}& RNN$_\text{EMA}^{\text{(AC)}}$    & 1.33       & 1.46        & 48.4\% & 280.2\\
    \hline
    \rule{0pt}{16pt}& RNN$_\text{EMA}^{\text{(LE, ST)}}$    & 1.29       & 1.36            & 61.3\%  & 220.4\\
    \hline
    \rule{0pt}{16pt}& RNN$_\text{EMA}^{\text{(LE, AC)}}$    & 1.29        & 1.39    & 54.8\% & 231.5\\
    \hline
    \rule{0pt}{16pt}& RNN$_\text{EMA}^{\text{(ST, AC)}}$    & 1.26        & 1.38        & 48.4\% & 210.5\\
    \hline
  \end{tabular}
\end{threeparttable}
\end{table}

There was no statistically significant difference among the RMSE$_\text{val}$ values of the models with different feature combinations ($p=0.119$, repeated measures ANOVA) (Table \ref{tab:feature}). The model using all features performed the best based on $<$1h and AIC. RNN$_\text{EMA}^{\text{(LE, ST)}}$ would be another good options for those who want to reduce the number of sensors based on RMSE$_\text{test}$ and $<$1h.

\subsection{Error Analysis}
\begin{figure}[tbp]
\centering
\centering
\includegraphics[width=0.9\linewidth]{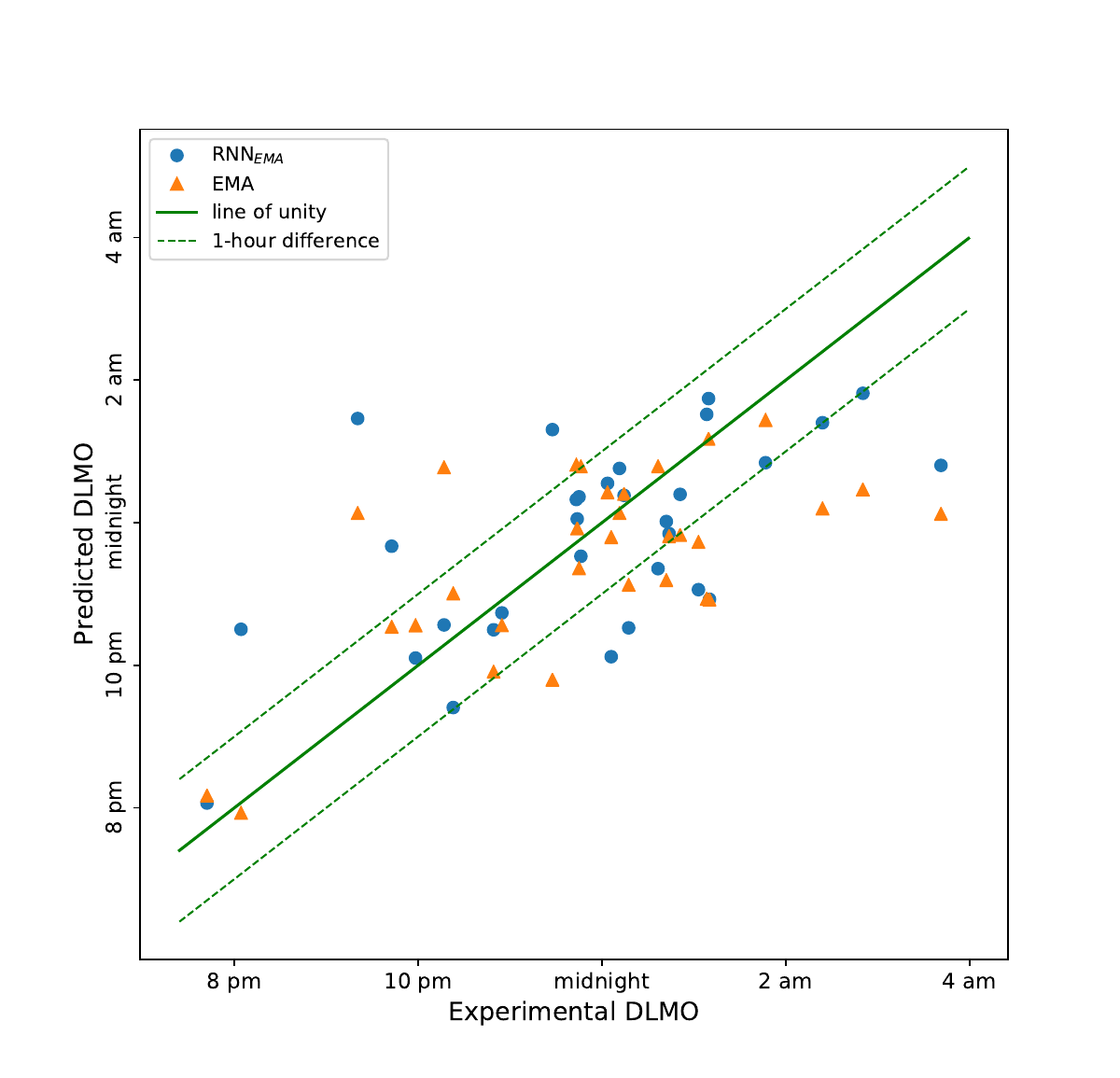}
\caption{The relationship between the predicted DLMO values from two methods and the experimental DLMO.}
\label{fig:error}
\end{figure}

For the relationship between the predicted DLMO of our EMA and RNN$_\text{EMA}$ models and true DLMO value for the test data (Figure \ref{fig:error}), the Pearson correlation of the prediction errors (i.e., absolute different between predicted DLMO and true DLMO) of two models was 0.596 with $p=4.0\times 10^{-4}$, which suggests a strong correlation between the errors. Visual inspection shows that the model tends to predict later DLMO if the experimental DLMO is prior to midnight, and predict earlier DLMO if the experimental DLMO is after midnight.

To understand how to improve EMA model, we studied whether missing patterns in sleep parameters and sleep regularity of a participant influence the performance of the model. We first split the data into two categories based on whether the absolute error is less than 1 hour or not. Then, we used Mann-Whitney U test to compare the missingness/sleep regularity in these two categories. There were no statistically significant relationships in (i) the relationship between the number of missing values in the sleep parameters and DLMO prediction errors of EMA and (ii) the relationship between standard deviations of sleep parameters across days (i.e., sleep regularity) and DLMO prediction errors of EMA.

\section{Discussion}
\subsection{Choice of Moving Average Models}

Previous work that analyzed the relationship between sleep data and DLMO value \citep{martin2002sleep,burgess2005dim,crowley2006estimating} used only SMA models. In this paper, we compared SMA with the other two moving average models. 

\paragraph{SMA v.s. EMA} In Section \ref{sec:hpt}, we hypothesized that the EMA model is more sensitive to noise but its performance is better when the noise is limited. Our experiment supported this (Figure \ref{fig:noise}): EMA performs better than SMA when the noise is small ($\sigma<2$) but it is worse when the noise is large ($\sigma>2$). With the development of technologies that reduce noise, an EMA model could be more useful in the future.\\
\paragraph{SMA/EMA v.s. MA} In this work, we collected 223 samples. According to the analysis in Section \ref{sec:hpt}, the sample size was large enough, but the prediction performance of MA was similar to that of SMA/EMA. As a result, MA was not preferred for predicting DLMO using sleep midpoints.

\subsection{The Performance of the Proposed Two-Step Models}

\begin{table}
\centering 
\begin{threeparttable}
  \caption{A comparison between RNN$_\text{EMA}$ and previous work using a subset of the data in this datasets. For the values that are not reported, we use the symbol '/'.}
  \label{tab:comp} 
  \begin{tabular}{|C{3.8cm}|C{2.1cm}|C{2.1cm}|C{1.6cm}|}
      \hline
    Model    & RMSE$_\text{val}$    & RMSE$_\text{test}$& $<$1h\\
    \hline
    RNN$_\text{EMA}$        & 1.21    & 1.38    & 64.5\%\\
    \hline
    Neural Network \cite{brown2018neural} & 1.45        & 1.49            & /    \\
    \hline
    Limit Cycle Oscillator \cite{phillips2017irregular}    & /        & /    & 43\% \\
    \hline
  \end{tabular}
\end{threeparttable}
\end{table}

In this paper, we proposed a two-step framework using two time-scale features for estimating DLMO. Our experiment showed that the performance of the two-step models was significantly better than the models using only daily sampled or frequently sampled data. We also compare our model with other work \citep{phillips2017irregular, brown2018neural} using a subset of the data in this datasets (Table \ref{tab:comp}). Our model performed better on the $<$1h metric (65\% v.s. 43\%) than a limit cycle oscillator based method with light input \citep{phillips2017irregular}. The performance of our model was also better than that of a neural network method \citep{brown2018neural} that used sleep timing, frequently sampled sensor data (LE and AC) and demographic information (gender, personality types and sleep quality index) as input features (RMSE$_\text{val}$: 1.21 v.s. 1.45, RMSE$_\text{test}$: 1.38 v.s. 1.49). Neither of those publications included the other metrics.

The error analysis showed that the error of our RNN$_\text{EMA}$ model was strongly correlated with the error of EMA. Therefore, one approach to improve RNN$_\text{EMA}$ is to improve the accuracy of the moving average model.

This novel framework is the generalization of current models that leverage both daily sampled data and frequently sampled data. The model proposed in this paper may be eventually useful for estimating DLMO as a marker of human circadian phase. Every feature we used in our model is derived from wearable sensors that are low-cost, easy to use, and can provide data in close to real time. As noted above, some existing methods \citep{kolodyazhniy2011estimation,kolodyazhniy2012improved,gil2013human,stone2019generalizability} were designed for estimating melatonin instead of predicting DLMO directly. These existing methods therefore require a complete melatonin profile (e.g., the shortest length of data was 22-hour long in these reports \citep{gil2013human}) for scaling the melatonin amount when training the model, which is expensive and inconvenient. The model proposed in the work directly estimates DLMO and requires much less melatonin collection for training the model.

The ideas of this work can be applied to other regression or classification tasks with sensor data with different sample rates. In this work, we have two different time scales of data: the daily sampled data capture behavioral or physiological parameters from each day, and the frequently sampled data are stored minute by minute. Other times scales of data can be also investigated.

\subsection{Limitation and Future Work}

There are some limitations in our work. First, we did not compare this model with previous work \citep{kolodyazhniy2011estimation,kolodyazhniy2012improved,gil2013human,stone2019generalizability} because those works developed models to estimate the amount of melatonin first and then derived DLMO based on the estimated melatonin profile, which required normalizing the melatonin data by fitting the curve using sufficient melatonin data. Since our melatonin was sampled from 13 hours long laboratory studies, it was not sufficient for fitting the curve. We leave this comparison to future work and we will also explore whether we can train a better model using all melatonin data, not just DLMO.

Second, the dataset we used were collected from students at a single college. The relationship of the parameters may be different in other groups of people. The models need to be tested in other populations for confirming the generalizability. We will further evaluate the performance of our models on other populations such as shift workers, people with jet lag, and people with psychiatric disorders to explore clinical application of our model.

Third, we explored the best feature combination for estimating DLMO. Our current result (Table \ref{tab:feature}) was not conclusive, as different metrics in our experiment produced different results for different feature combinations. The comparison of feature combination has to be further investigated with a larger dataset.

As additional future work, we will explore whether the model can be improved by utilizing other features including phone screen time and caffeine intake, which also affect human circadian phase \citep{cajochen2011evening,burke2015effects}, as well as specifics of physical/mental condition. Non-linear models will be examined for sleep parameters, and we will compare this framework with other existing approaches. Potential applications using sensor-based DLMO estimation include adjusting the timing of drug administration and designing better work schedules for each individual based on their circadian phase; these and other applications requiring knowledge of circadian phase will be more feasible once non-invasive inexpensive long-term monitoring methods are available.

\section{Conclusion}
In this paper, we proposed and evaluated a two-step framework that leverages features of two time scales for predicting DLMO. This framework is the extension of previously used models with input of either frequently sampled data (i.e., light exposure, skin temperature, physical activity) or daily sampled data (i.e., sleep onset/offset timing). The experiment showed that the proposed model performed better than the model only using one-time-scale feature.

\begin{acks}
We thank research support by NSF \#1840167, NIH R01GM105018, F32DK107146, T32HL007901, K24HL105664, R01HL114088, R01HL128538, P01AG009975, and R00HL119618, Samsung Electronics, and NEC Corporation.
\end{acks}
\bibliographystyle{ACM-Reference-Format}
\bibliography{dlmo}

\end{document}